\documentclass[11pt]{article}

\usepackage[preprint]{acl}

\usepackage{times}
\usepackage{latexsym}

\usepackage[T1]{fontenc}
\usepackage[utf8]{inputenc}

\usepackage{microtype}

\usepackage{inconsolata}

\usepackage{graphicx}

\usepackage{amsmath}

\title{Structured yet Bounded Temporal Understanding in Large Language Models}

\author{Damin Zhang \\
  Purdue University \\
  \texttt{zhan4060@purdue.edu} \\\And
  Julia Rayz \\
  Purdue University \\
  \texttt{jtaylor1@purdue.edu} \\}

\begin{document}
\maketitle
\begin{abstract}
Large language models (LLMs) increasingly show strong performance on temporally grounded tasks, such as timeline construction, temporal question answering, and event ordering. However, it remains unclear how their behavior depends on the way time is anchored in language. In this work, we study LLMs' temporal understanding through temporal frames of reference (t-FoRs), contrasting deictic framing (past-present-future) and sequential framing (before-after). Using a large-scale dataset of real-world events from Wikidata and similarity judgement task, we examine how LLMs' outputs vary with temporal distance, interval relations, and event duration. Our results show that LLMs systematically adapt to both t-FoRs, but the resulting similarity patterns differ significantly. Under deictic t-FoR, the similarity judgement scores form graded and asymmetric structures centered on the present, with sharper decline for future events and higher variance in the past. Under sequential t-FoR, similarity becomes strongly negative once events are temporally separated. Temporal judgements are also shaped by interval algebra and duration, with instability concentrated in overlap- and containment-based relations, and duration influencing only past events under deictic t-FoR. Overall, these findings characterize how LLMs organize temporal representation under different reference structures and identify the factors that most strongly shape their temporal understanding.
\end{abstract}

\section{Introduction}
Temporal understanding is a fundamental component of human intelligence, allowing us to represent, interpret, and reason about time and how events unfold in relation to it. Large language models (LLMs) have demonstrated strong performance on natural language understanding tasks, including temporally grounded tasks, such as temporal question answering \cite{zhou2019going,gurnee2023language,islakoglu2025chronosense}, timeline construction \cite{nylund2024time,bazaga2025learning}, and event prediction \cite{dhingra2022time}, and narrative understanding \cite{min2023recent}. However, temporal understanding still remains a challenge to LLMs \cite{zhou2019going,jain2023language,wallat2024temporal,qiu2024large}.

Various notions of time have been used in temporal understanding in the realm of natural language understanding. Some approaches adopt an absolute time where events are anchored to normalized timestamps and evaluate whether the systems are able to answer temporally grounded questions. Other works adopt a more relational time that models the relations between events. A mixed notion of time combines both views that align events to an absolute time and reason over their relative order. Among these notions, time is treated as a structural resource for ordering and constraining events.

From a different perspective, linguistic studies show that time is encoded through tense, aspect, temporal adverbials, and discourse structure, and cognitive science works have argued that the representation of time is metaphorical of space \cite{mcglone1998back,lakoff1999review,gentner2002time}. As LLMs are trained on natural language, the data does not reflect the raw physical time, but linguistically and cognitively shaped representations of time. From this perspective, LLMs inherit the temporal abstractions encoded in natural language, including spatiotemporal representation \cite{gurnee2023language} and temporal intervention \cite{nylund2024time}. Therefore, the appropriate question is more about how well LLMs capture the kind of temporality natural language itself encodes.

As human beings possess the ability to dynamically reason about temporal relations among events relative to a reference point under frames, we draw on the distinction between two temporal frames of reference (t-FoRs): deictic and sequential t-FoRs \cite{evans2013temporal}. A t-FoR specifies how temporal relations are organized relative to a reference point. For example, in deictic t-FoR, the reference point is anchored in the experiencer's ``now", or the ``ego" in other terms, allowing the notions of past/present/future; in sequential t-FoR, it is anchored in a specific event to create the before/after notions. In this work, we attempt to explore LLMs' temporal understanding through the lens of t-FoRs. By manipulating reference point in each t-FoR, we can probe how LLMs adapt their temporal understanding and where their representations exhibit uncertainty. Inspired by recent work on language models' temporal cognition\cite{li2025other}, we incorporate similarity judgement task to observe the induced patterns. We aim to answer the following questions:
\begin{itemize}
    \item \textbf{RQ1}: Do large language models adapt to different temporal frames of reference?
    \item \textbf{RQ2}: What factors affect the temporal understanding in large language models?
\end{itemize}

Our contributions can be summarized as follows:
\begin{itemize}
    \item We introduce a frame-sensitive evaluation framework for temporal understanding in large language models, contrasting deictic and sequential t-FoRs through a similarity judgement task applied to real-world events.
    \item We propose a large-scale temporal dataset from Wikidata that preserves interval structure and duration, enabling systematic analysis of temporal distance, Allen's interval algebra, and event duration at scale.
    \item We provide comprehensive empirical evidence that LLMs' temporal behaviors are strongly frame-dependent and shaped by structural temporal factors, including past-future asymmetry under deictic t-FoR, rapid similarity collapse under sequential t-FoR, elevated variability for overlap-based relations, and duration effects retricted to past events.
\end{itemize}

\section{Related Work}
Research on temporal understanding in NLP has made substantial progress, but much of it implicitly targets specific levels of the above categories. A large body of works focuses on tasks such as temporal expression normalization \cite{bethard2007cu,strotgen2010heideltime}, event ordering \cite{chambers2008unsupervised,ning2018improving,ning2018multi}, timeline construction \cite{do2012joint}, and temporal question answering \cite{zhou2019going,chen2021timeqa}. These tasks primarily evaluate a model's ability to represent and reason about temporal relations among events - that is, how events are positioned with respect to one another in time. More recent work explores time-aware embeddings, temporal pretraining objectives, and structured temporal representations, further strengthening this capability.

\subsection{Temporal Reasoning Benchmarks}
Question-answering datasets such as SQuAD \cite{rajpurkar2016squad,rajpurkar2018know} and Natural Questions \cite{kwiatkowski2019natural} construct temporal questions within a single, fixed time period. However, some questions may have different answers depending on when they are asked. To address this, several datasets include time-sensitive questions, such as TimeQA \cite{chen2021timeqa}, TempLAMA \cite{dhingra2022time}, SituatedQA \cite{zhang2021situatedqa}, StreamingQA \cite{liska2022streamingqa}, RealtimeQA \cite{kasai2023realtime}, TempReason \cite{tan2023towards}, and MenatQA \cite{wei2023menatqa}.

Temporal commonsense knowledge, such as a fact that a leap year generally occurs every four years, is another important aspect of temporal reasoning. Datasets like MCTACO \cite{zhou2019going} and TimeDial \cite{qin2021timedial} focus specifically on evaluating models' understanding of temporal commonsense.

Beyond datasets targeting specific temporal properties or sequential relations, comprehensive benchmarks such as TRAM \cite{wang2024tram} and TimeBench \cite{chu2024timebench} evaluate temporal reasoning across multiple tasks. Additionally, datasets such as TimeQuestions \cite{jia2021complex}, ExpTime \cite{yuan2024back}, TGQA \cite{xiong2024large}, and ToT \cite{fatemi2024test} leverage knowledge graphs to assess models' understanding of time-event and event-event relations. Datasets such as ChronoSense \cite{islakoglu2025chronosense} utilized Allen's interval algebra \cite{allen1983maintaining} to assess LLMs' temporal understanding capability.

\subsection{Frame of Reference}
A frame of reference (FoR) is a coordinate system used to determine the position of a figure relative to a ground from a particular perspective \cite{talmy2003toward}, typically corresponding to the viewpoint of an observer.

From a philosophical perspective, time can be conceptualized as either an A-series or a B-series \cite{mctaggart1908unreality}. The A-series represents a tensed and dynamic view of time, where events are located relative to the observer's subjective ``now'' (the deictic center) as past, present, or future. In this view, a future event successively becomes present and eventually shifts into the past. In contrast, the B-series reflects a tenseless and static ordering of events, where each event is positioned as earlier than or later than another, independent of any observer's temporal standpoint.

From a cognitive perspective, time is commonly described using two conceptual metaphors: moving ego (ME) and moving time (MT) \cite{mcglone1998back,boroditsky2000metaphoric,gentner2002time}. In the ME perspective, time is stationary while the observer (ego) moves forward through it, such as ``We are approaching Tuesday''. In the MT perspective, the observer remains stationary while time itself flows toward them, for example, ``Tuesday is approaching''.

A temporal frame of reference (t-FoR) typically involves three components: a target event (TE), a reference point (RP), and an origo (O). The TE denotes the event being fixed, the RP provides the point of comparison relative to the TE, and the O serves as the anchoring perspective. As on the of the first attempts to integrate temporal perspectives within a FoR framework, \citeauthor{moore2004ego} distinguishes between ego-centered MT and non-deictic MT, proposing an ego-based vs. field-based taxonomy to capture these differences. Extending this approach, \citeauthor{moore2011ego} groups ME and ego-centered MT as ``ego-perspective'', corresponding to the A-series, while field-based perspectives correspond to the B-series. The reference-point metaphors framework splits the MT perspective into two categories: ego-based and field-based, where former merges with the ME perspective under a deictic clasification \cite{nunez2006future}. The temporal framework models taxonomy integrates reference-point metaphors with descriptions of time across all three spatial frames of reference (s-FoRs) \cite{kranjec2006extending}. Building on these developments, \citeauthor{evans2013temporal} aligns the A-series with ego-based frames \cite{moore2004ego} and ego-RP metaphors \cite{nunez2006future}, and the B-series with field-based frames \cite{moore2004ego} and time-RP metaphors \cite{nunez2006future}. Other revisions classify ME and MT as ego-relative temporal motion constructions, while sequence-based conceptions are termed positional time constructions \cite{sinha2011time}. \citeauthor{tenbrink2011reference} characterizes time as possessing an ``inbuilt asymmetry'', which can be conceptualized either as a vector from past to future or as a vector expressing anteriority/posteriority in event sequences.

\subsection{Temporal Reasoning in Language Models}
Although Large Language Models (LLMs) have demonstrated impressive reasoning capabilities through in-context learning and post-training techniques, understanding and reasoning about temporal information remains challenging \cite{jain2023language}. Recent studies have highlighted that LLMs particularly struggle with relative time references, randomized time references, and typical temporal questions \cite{wallat2024temporal}.

Prior work has sought to enhance LMs' temporal understanding through several approaches, including post-training with external knowledge \cite{yuan2024back,tan2024towards,xiong2024large}, leveraging code-format representations \cite{li2023unlocking,zhu2023question}, and improving generalization across diverse temporal tasks \cite{su2024timo}.

However, these approaches leave two issues underspecified. First, the frame of reference governing temporal judgements is frequently implicit, making it difficult to distinguish deictic from sequential reasoning. Second, temporal relations are often treated symmetrically or discretely, obscuring graded uncertainty and relation-specific difficulty.

\section{What Temporal Understanding Means for Language Models}
It is necessary to clarify what is reasonable to expect from language models (LMs) in terms of temporal understanding. In this work, we use the term `event" as a temporally bounded occurrence that can be described and related to other occurrences. Our goal is to study how language models organize such events in time. 

LMs do not encounter time as human beings do, but only through what is encoded in natural language. Therefore, whatever temporal understanding LMs may acquire is necessarily mediated by linguistic and cognitive representation of time that is present in human discourse. This makes the temporal structure available to LMs being perspectival, different from physical or metaphysical notions of time, that LLMs do not have access to absolute time nor do they experience lived temporality.

This places LLMs in an intermediate place between the philosophical and cognitive views. On one hand, LLMs model temporal structure as stable relations between events, resembling what McTaggart \cite{mctaggart1908unreality} classified as B-series structure: the ``earlier than" and ``later than" relations between events, as prior works have shown \cite{xiong2024large}. On the other hand, natural language also encodes so called A-series structure that distinguishes positions from past to present to future, relative to an observer's viewpoint. These perspectival distinctions arise from how the model is situated within it. Though prior works have shown success on temporally grounded tasks, but they rarely test whether LLMs can adopt and update temporal perspectives.

Temporal frames of reference (t-FoRs) provide a way to probe this capacity. In a deictic t-FoR, temporal relations are anchored through experiencer-based reference strategy, allowing distinctions between past, present, and future. In a sequential t-FoR, the reference strategy is changed to event-based, where the reference point is anther event, creating before and after relations. Both frames are pervasive in language, linguistically grounded, and cognitively motivated \cite{evans2013temporal}.

\begin{figure}
    \centering
    \includegraphics[width=1.0\linewidth]{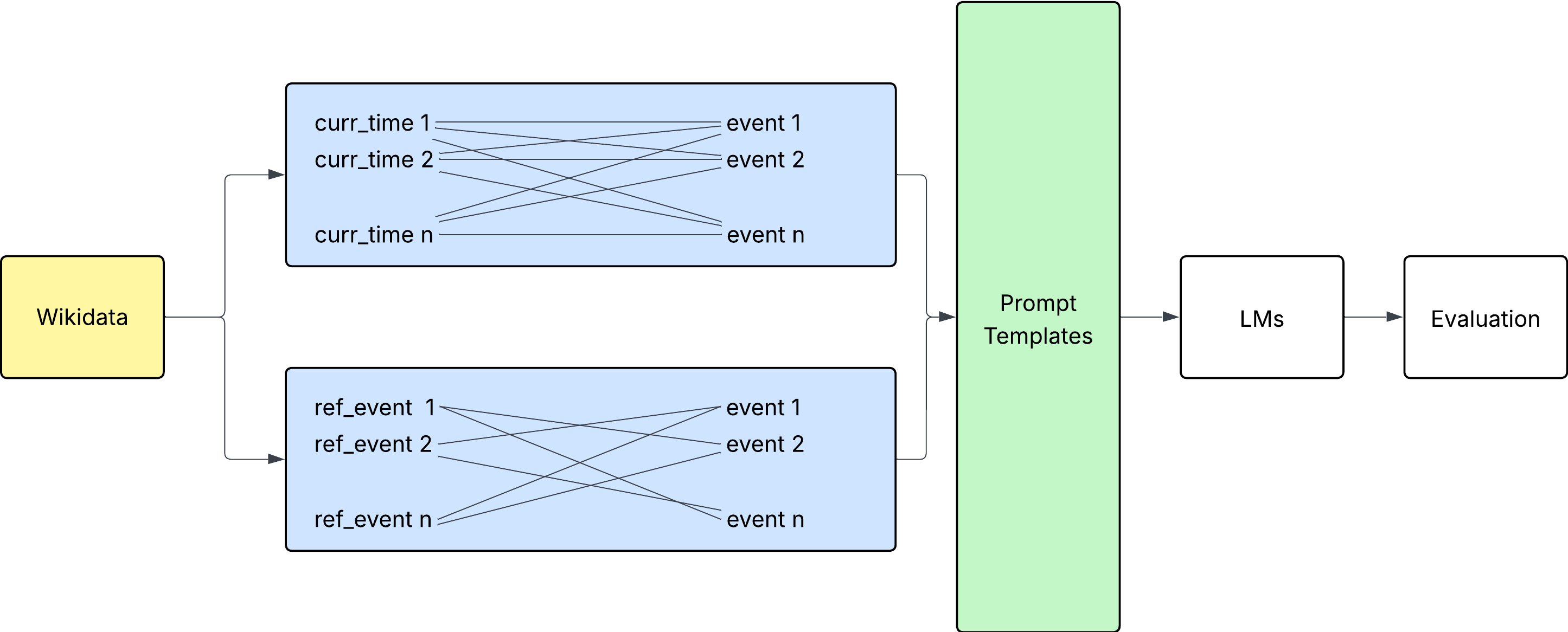}
    \caption{Overall benchmark steps.}
    \label{fig:overall}
\end{figure}
We propose a frame-dependent benchmark that aims to probe temporal understanding in LLMs. Intead of evaluating factual correctness from multiple choices format, we measure graded similarity judgements as a function of temporal relations, temporal distance, and reference-point manipulation. As illustrated in Figure \ref{fig:overall}, the benchmark adopts an experiencer-based reference strategy from the deictic t-FoR and event-based strategy from the sequential t-FoR.

\subsection{Event Extraction}
We constructed a larger scale dataset by extracting real events from Wikidata \cite{vrandevcic2014wikidata}. The extracted finite set of events $E={e_1, e_2, ..., e_n}$, where each event $e_i=(t^{start}_i, t^{end}_i, l_i)$ is represented by a start time, end time, and name. We end up with 49,956 events dated from ``0001-01-01" to ``2100-12-31".

To avoid that models rely on memorized knowledge of specific historical events, we anonymized all event names using a list of rare animal names, while keeping start and end timestamps unchanged. The mapping from original to anonymized names is stored for prompt construction. Additionally, because events are temporally extended rather than instantaneous, we leveraged Allen's Interval Algebra \cite{allen1983maintaining} to characterize the full range of possible relations between two events. For later analysis and for balanced coverage across relations, we randomly sample 321 anonymized events such that all 13 Allen relations are represented.

\subsection{Temporal Frames of Reference}
We evaluate temporal understanding under two distinct temporal frames of reference (t-FoR), following Evan's taxonomy \cite{evans2013temporal}.

In the deictic t-FoR, temporal relations are established relative to a reference time $\tau \in T$, which is treated as the experiencer's ``now". A sequence of reference times of ``now", $T_{now}={\tau_1, \tau_2, ..., \tau_m}$, is created by enumerating from the earliest event start time to the latest event end time. Therefore, the data pairs are constructed as $(\tau,e_i)$ for $\tau \in T$ and $e_i \in E$.

In the sequential t-FoR, the reference point is not an external ``now" but an event itself. For each event $e_j \in E$, the temporal relations between $e_j$, the reference event, and all other events $e_i$ are computed. The resulted data pairs are $(e_i,e_j)$ for $e_i, e_j \in E$ and $e_i \neq e_j$.

For each data pair, both temporal relation and distance are calculated. The deictic t-FoR yields past/present/future relations and before/after for sequential t-FoR. The temporal distance is calculated in the unit of a day:
\begin{equation*}
    \Delta(r, e_i)=
    \begin{cases}
        t^{end}_i - t^{start}_r, & \text{if}\ t^{start}_r >= t^{end}_i \\
        t^{start}_i - t^{end}_r, & \text{otherwise}
    \end{cases}
\end{equation*}
where $r$ stands for the reference time/event and $e_i$ refers to the target event. The temporal distance is signed that negative values correspond to past/before relations; 0 indicates present relation; and positive values means future/after relations.

\begin{figure}
    \centering
    \includegraphics[width=1.0\linewidth]{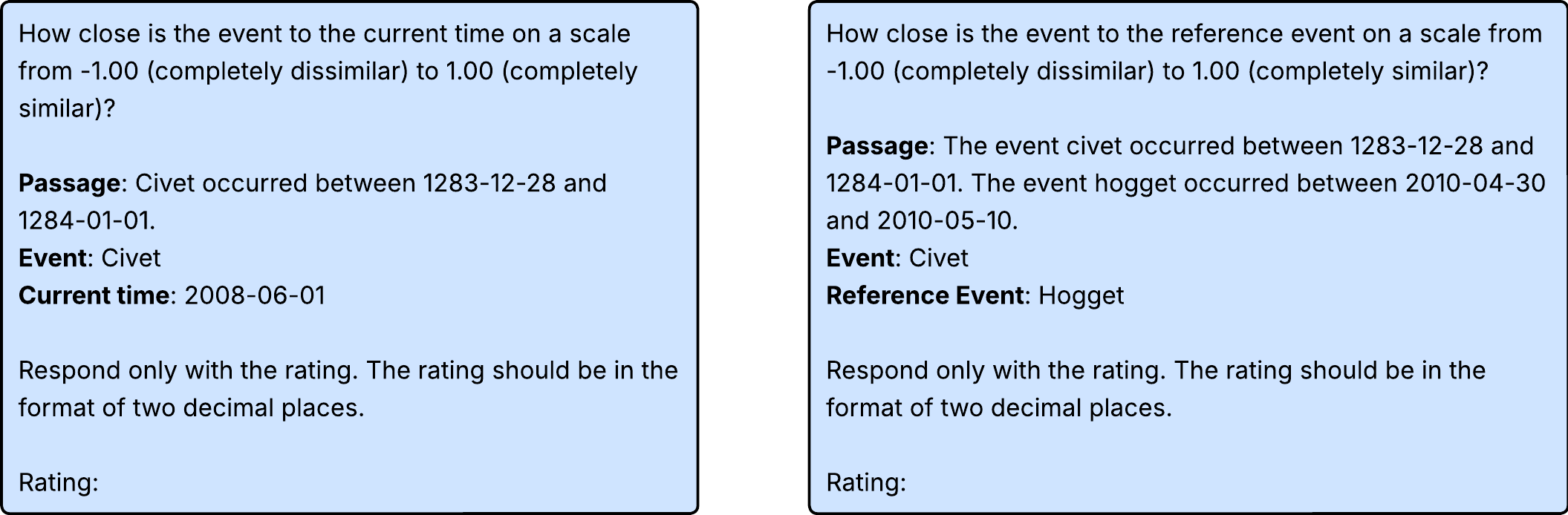}
    \caption{Prompt examples for similarity judgement. Left prompt is for deictic temporal frame of reference. Right prompt is for sequential temporal frame of reference.}
    \label{fig:prompts}
\end{figure}

\subsection{Similarity Judgement}
Inspired by recent works \cite{marjieh2022predicting,li2025other}, we leveraged similarity judgement task to probe LLMs to rate how similar the reference time/event is to the target event. Specifically, for each data pair $(r, e_i)$, where $r$ is the reference time/event, the tested LLMs are prompted to output a similarity score:
\begin{equation*}
    s_{LLM}(r, e_i)\in [-1.00, 1.00]
\end{equation*}
where 1.00 indicates mostly similar and -1.00 is completely dissimilar. The two prompt templates are shown in Figure \ref{fig:prompts}.

Overall, the dataset contains 2,916,927 data entries for deictic t-FoR and 51,360 data entried for sequential t-FoR.

\section{Experiment Setup}
In this work, we focused on evaluating two open-source large language models (LLMs) with different sizes: Qwen3-4B (Qwen3-4b-instruct) and Qwen3-30B (Qwen3-30b-a3b-instruct-2507) \cite{qwen3technicalreport}. The models can generate at most 64 new tokens for answers and scores are extracted using regex rules.

\subsection{Measurement}
For each signed temporal distance $\Delta=i$, the mean similarity score is computed by aggregating all outputs associated with that distance:
\begin{equation*}
    \overline{s}^{i} = \frac{1}{n_i}{\sum_{j=1}^{n_i} s_{LLM}^{(j)}}
\end{equation*}
where $n_i$ is the total number of data pairs whose temporal distance equals $i$.

Visualizing the resulting mean similarity values as a function of temporal distance yields a characteristic response pattern that reflects the model's temporal cognition. Under a deictic t-FoR, it is expected to manifest as a maximum around the present reference point, with similarity gradually decresing as events move further into the past or future. Deviations from this pattern indicate asymmetries or instability in how the model interprets temporal relations as the deictic anchor shifts.

Under a sequential t-FoR, a different pattern is expected as the before/after relations are symmetric with respct to temporal ordering, so that similarity scores should roughly result in a symmetric distribution centered around zero temporal distance.

\subsection{Allen’s Interval Algebra Analysis}
In addition to signed temporal distance, we further analyze LLMs' behavior patterns by leveraging Allen's Interval Algebra \cite{allen1983maintaining}, which defines 13 mutually exclusive temporal interval relations between two intervals: \textit{equals}, \textit{before}, \textit{after}, \textit{overlaps}, \textit{overlapped-by}, \textit{contains}, \textit{during}, \textit{started-by}, \textit{starts}, \textit{finished-by}, \textit{finishes}, \textit{meets}, and \textit{met-by}.

For each data pair, we computed the interval relations using the corresponding start and end times, allowing us to investigate whether LLMs certain interval relation introduce higher representational uncertainty. During evaluation, we grouped similarity scores by interval relation and computed the variance within each group. Higher variance indicates that the model has less consistent temporal assessments for that specific interval relation, even when temporal distance is comparable. This analysis allows us to identify where LLM judgements are most and least stable across the full structural space.

\section{Results}

\begin{figure*}[t]
    \includegraphics[width=0.48\linewidth]{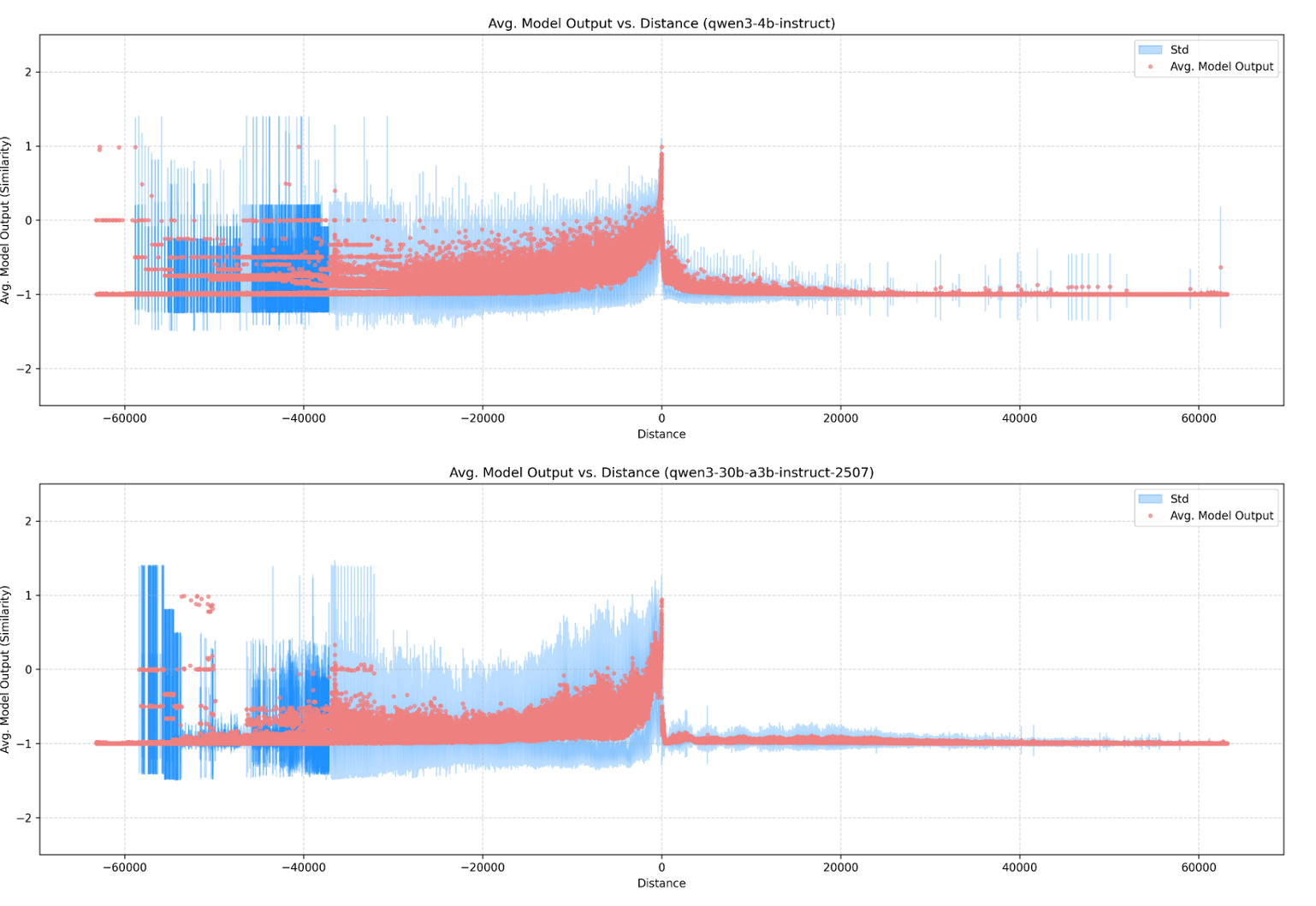} \hfill
    \includegraphics[width=0.48\linewidth]{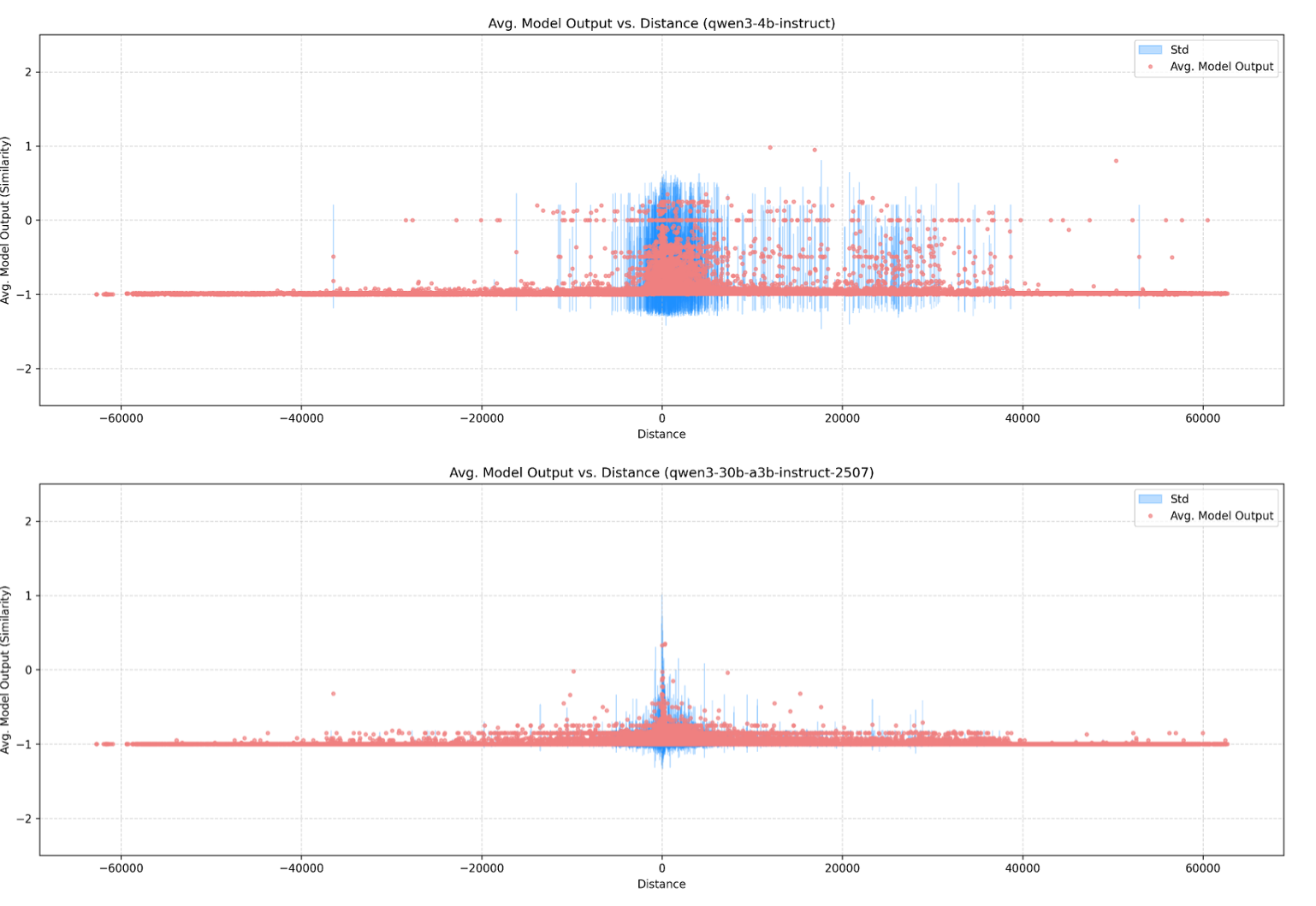}
    \caption{Mean similarity ratings vs. temporal distance. \textbf{Left}: in deictic t-FoR. \textbf{Right}: in sequential t-FoR. \textbf{Top}: qwen3-4b. \textbf{Bottom}: qwen3-30b. The blue shades indicate the variance of the ratings. Red dots correspond to mean similarity rating at a specific temporal distance.}
    \label{fig:rated}
\end{figure*}

\subsection{Function of Temporal Distance}
Figure \ref{fig:rated} shows the mean similarity score (red dots) against signed temporal distance for each data pair, together with the standard deviation across instances at the same distance (blue vertical bars). The left two plots correspond to the deictic t-FoR, and the right two plots correspond to the sequential t-FoR.

Under the deictic t-FoR, we observe a clear asymmetric distance effect for both LLMs. Similarity ratings peak near zero temporal distance and drop sharply for future events (positive distances), where scores rapidly converge to -1.00 with comparatively low variance. In contrast, past events (negative distances) exhibit a more gradual decline in similarity, accompanied by substantially higher variance, especially for events occurring tens of thousands of days before the reference point. This indicates that the models' treatment of the future is more uniform and compressed, whereas their representation of the past is richer but less stable. The pattern is more pronounced in the Qwen3-4b model, while the 30B model shows reduced variance near the present but preserves the same overall asymmetry.

Under the sequential t-FoR, the response curves differ significantly. Because the temporal relations are computed using event-based reference strategy, similarity scores remain strongly polarized for most non-zero distances, with values quickly saturating near the lower similarity bound. Variance is highest only in a narrow band around zero temporal distance: at temporal distance $\Delta=0$ this reflects the coexistence of multiple interval relations, such as \textit{overlaps}, \textit{meets}, or \textit{during}, while small non-zero distances still yield weakly separated \textit{before}/\textit{after} cases. Beyond this region, similarity ratings are highly stable and strongly negative, indicating that models treat most non-overlapping event pairs as mostly dissimilar in the sequential setting.

\begin{figure*}[t]
    \includegraphics[width=0.48\linewidth]{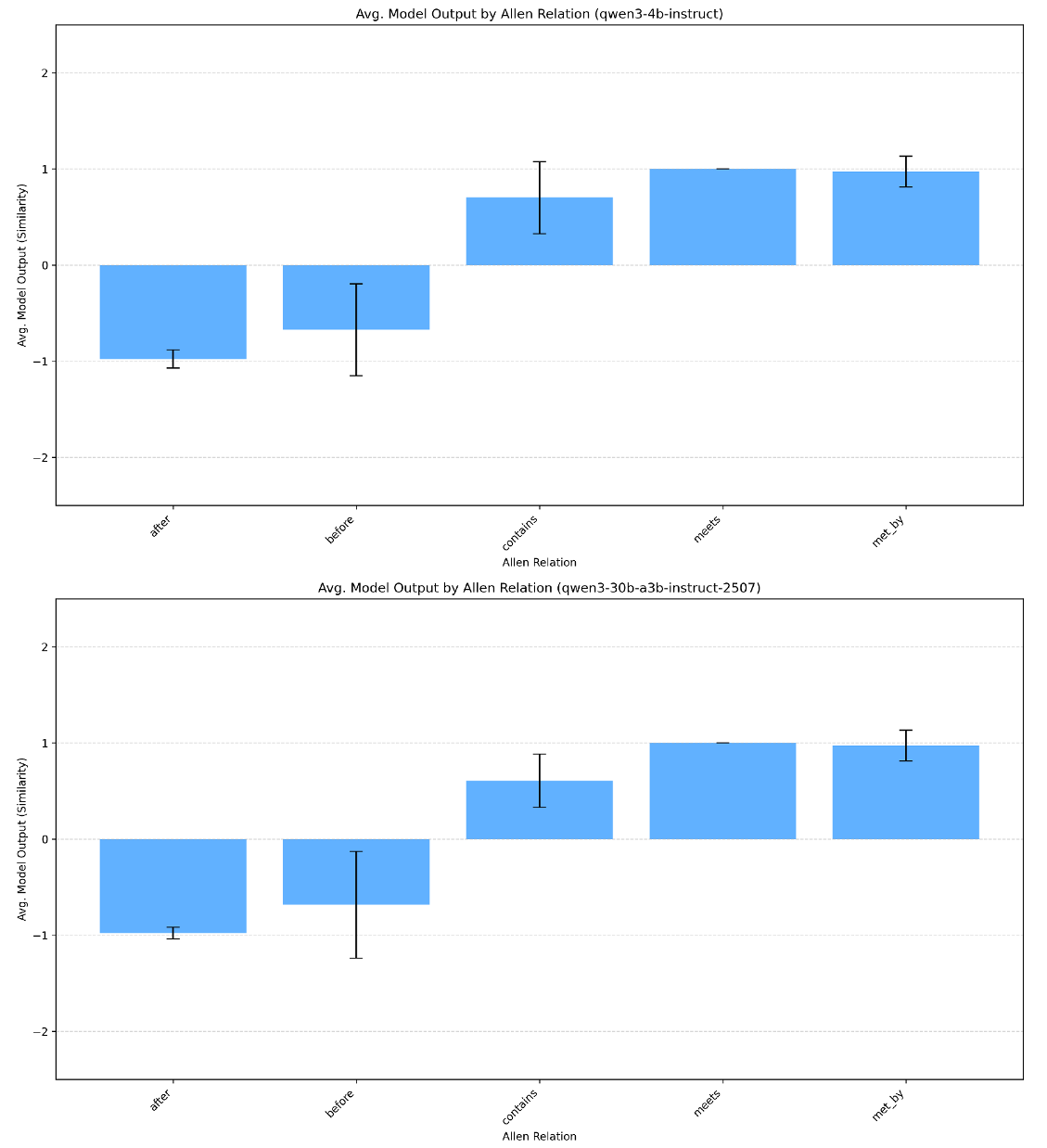} \hfill
    \includegraphics[width=0.48\linewidth]{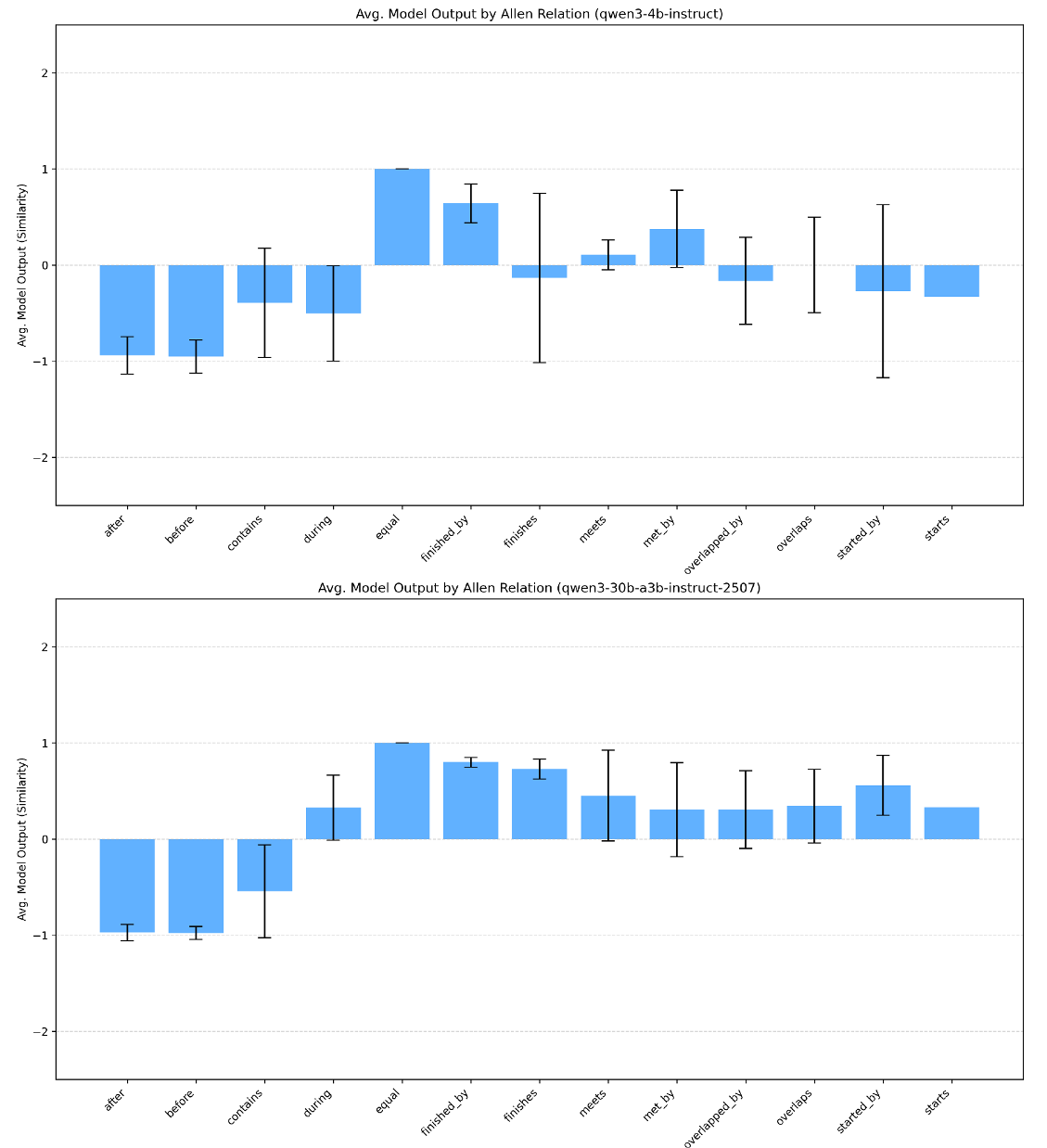}
    \caption{Mean similarity ratings by Allen's Interval Algebra. \textbf{Left}: in deictic t-FoR. \textbf{Right}: in sequential t-FoR. \textbf{Top}: qwen3-4b. \textbf{Bottom}: qwen3-30b. Bars indicate mean similarity score for each relation, and error bars denote the standard deviation across instances.}
    \label{fig:allen}
\end{figure*}

\subsection{Interval-Specific Variance}
Additionally, we leveraged Allen's Interval Algebra to better understand LLMs' behavior in special cases, as temporal distance $\Delta=0$ contains different interval realtions, for example, ``overlaps" or ``meets". We reported the mean similarity score and standard deviation for each relation type, as shown in Figure \ref{fig:allen}. The left two plots correspond to the deictic t-FoR, while the right two plots are in sequential t-FoR.

Across both LLMs and both t-FoRs, we observe strong polarity effects for interval relations that are strictly ordered: intervals labeled as \textit{after} consistently receive similarity ratings close to -1.00, followed by \textit{before}, indicating that events occurring entirely after or entirely before the reference event are treated as mostly dissimilar. Oppositely, interval relations implying coincidence or direct boundary contact yield the highest positive similarity scores, reflecting strong alignment when events coincide or touch at their endpoints, particularly in \textit{equal}, \textit{meets}, and \textit{met-by}.

More nuanced behavior is observed for containment and partial-overlap relations, such as \textit{contains}, \textit{during}, \textit{overlaps}, and \textit{overlapped-by}. These interval relations tend to have mid-range similarity scores but substantially higher variance, indicating less stable temporal assessments when interval topology is more complex. This variability is more pronounced under the sequential t-FoR, where temporal judgements depend on event-based reference strategy. Under the deictic t-FoR, variance is still elevated for partial-overlap relations, but similarity ratings remain more polarized, consistent with previously observed asymmetry between past and future.

The pattern holds across model sizes, although the Qwen3-30b model generally exhibits reduced dispersion relative to the 4B model, suggesting improved but still imperfect discrimination of structurally ambiguous temporal relations at larger scale.

\begin{figure*}[t]
    \includegraphics[width=0.48\linewidth]{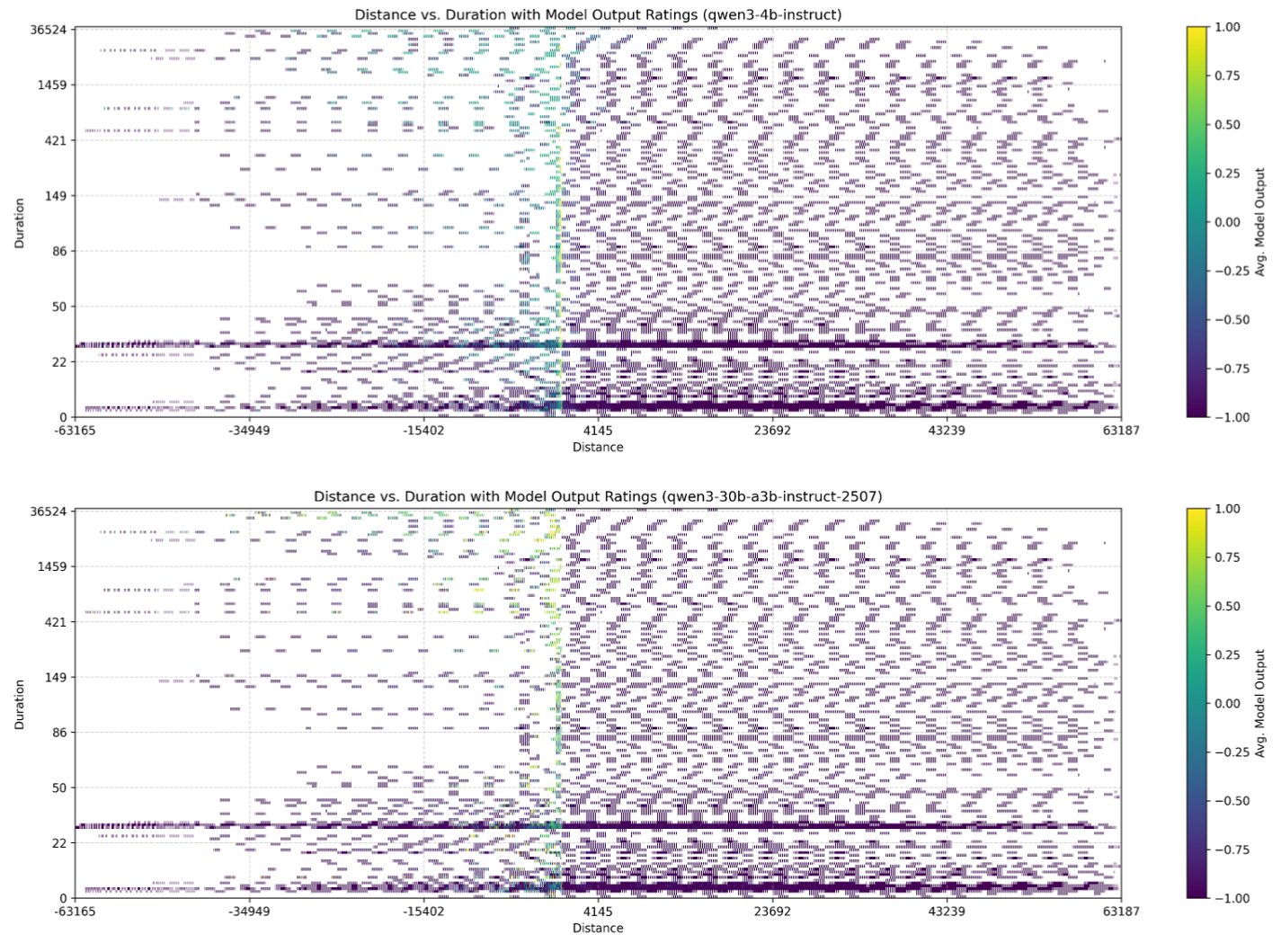} \hfill
    \includegraphics[width=0.48\linewidth]{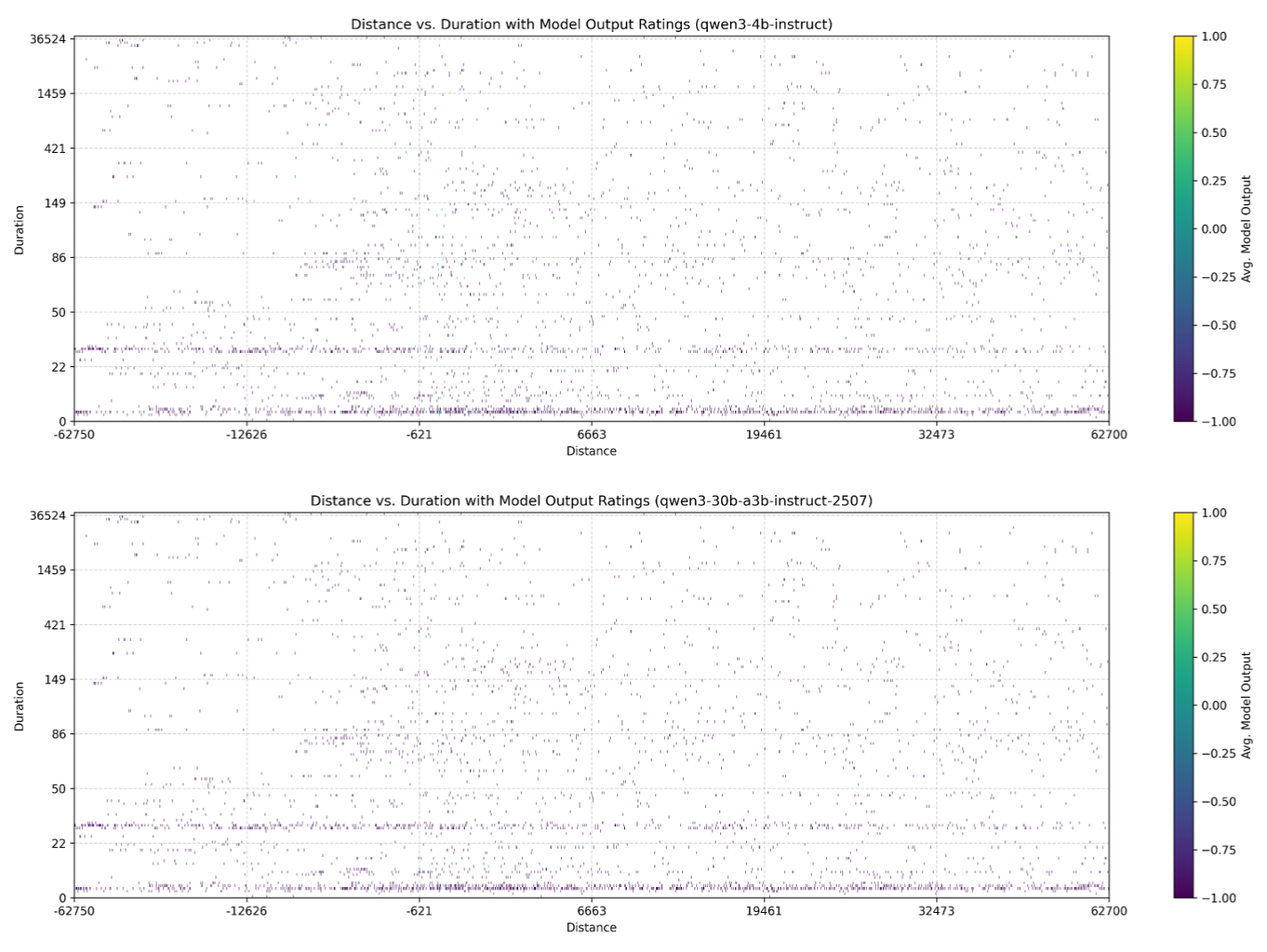}
    \caption{Heatmap of mean similarity ratings vs. (distance, duration). \textbf{Left}: in deictic t-FoR. \textbf{Right}: in sequential t-FoR. \textbf{Top}: qwen3-4b. \textbf{Bottom}: qwen3-30b.}
    \label{fig:drr}
\end{figure*}

\subsection{Interaction between Temporal Distance and Event Duration}
We further analyze whether temporal similarity depends jointly on temporal distance and event duration by averaging model outpus over (distance, duration) bins. Figure \ref{fig:drr} shows the results for both LLMs.

Under the deictic t-FoR, a clear pattern emerges for past events, that similarity increases with duration. Long-duration past events receive higher similarity ratings than short-duration ones at comparable distances, and in many cases distant long-duration past events are rated as more similar than short-duration events occurring much closer to the reference point. No comparable duration effect appears for future events, whose ratings rapidly converge to strongly negative values regardless of duration. The effect is qualitatively consistent across model sizes.

On the other hand, no systematic duration-distance structure is visible under the sequential t-FoR. Similarity ratings remain strongly polarized once events are temporally separated, and duration does not meaningfully modulate similarity.

\section{Discussion}
\subsection{Adaptation to Temporal Frames of Reference}
Based on the results, we observe that model behavior depends strongly on the choice of temporal frame of reference (t-FoR). Under the deictic t-FoR, similarity ratings vary smoothly with temporal distance and show a graded structure centered on the reference point. In contrast, under sequential t-FoR, similarity stabilizes into strongly negative values once events are temporally separated, with variability concentrated only near zero distance. This frame-specific divergence demonstrates that temporal understanding in large language models (LLMs) is frame-dependent, that they adapt their temporal judgements to the way time is anchored, whether to the experiencer's ``now" or to another event. This adaptation is systematic rather than random, suggesting that temporal framing is encoded at a representational level, not only at the surface form of the task.

\subsection{Effects of Temporal Distance and Interval Structure}
The Allen's interval algebra result shows that similarity judgements are most stable for strictly ordered relations (\textit{before}/\textit{after}) and boundary-aligned or coincident relations (\textit{meets}/\textit{met-by}/\textit{equal}), and least stable for overlap- and containment-based relations. This reveals that temporal understanding in LLMs is not merely a function of linear distance, but also of how intervals relate structurally.

The reduced stability for overlap relations suggests that LLMs handle ambiguous intervals less consistently than cleanly ordered ones. This aligns with the higher variance observed at small temporal distances, where multiple Allen's interval relations remain possible. These findings indicate that interval geometry is a key determinant of model behavior, that temporal understanding is sensitive to the structural complexity of temporal relations.

\subsection{Effects of Event Duration and Past-Future Asymmetry}
A second factor shaping temporal understanding is event duration, but only under specific conditions. Under the deictic t-FoR, duration systematically modulates similarity for past events: long-duration past events are judged as more similar than short-duration ones at comparable distances, and even more similar than short-duration events that occur much closer to the reference point. Crucially, the same effect does not appear for future events or under sequential t-FoR.

This selective retrospective duration sensitivity suggests that time is not represented symmetrically in LLMs. The past is encoded in a richer and more graded structure, while the future collapses toward uniform low similarity. The fact that duration effects disappear under sequential t-FoR reinforces that this is not purely a semantic property of events themselves, but a property of how LLMs anchor temporal meaning.

Therefore, temporal understanding in LLMs is influenced not only by temporal distance and interval structure, but also by duration. These effects are frame-dependent and directionally asymmetric.

\subsection{Summary of Findings}
Taken together, these findings indicate that LLMs demonstrate structured but bounded temporal understanding. They respond coherently to different frames of reference and their behavior is systematically modulated by temporal distance, interval relations, and duration. However, these effects are not uniform: temporal representations are richer in the past than in the future, and more stable for cleanly ordered relations than for overlapping intervals.

This suggests that LLMs' temporal understanding is shaped by linguistically acquired regularities rather than explicit temporal inference mechanisms. Temporal representation appears to be encoded as a graded, context-anchored similarity space, that adapts to framing, but does not collapse into a single global temporal metric.

\section{Conclusion}
We introduced a frame-sensitive evaluation of temporal understanding in large language models (LLMs), comparing behavior under deictic and sequential temporal frames of reference (t-FoRs). Our results show that LLMs adapt systematically to the chosen t-FoR: under deictic t-FoR, similarity judgements form graded and asymmetric patterns centered on the present, whereas under sequential t-FoR, similarity rapidly collapses to strongly negative values once events are temporally separated. Temporal judgements are additionally shaped by interval structure and duration, with instability concentrated in overlap-based relations and with duration affecting only past events under deictic anchoring. Altogether, these findings indicate that LLMs exhibit a structured yet bounded form of temporal understanding.

\section{Limitations}
This work currently focuses only on English text, leaving extension to other languages for future explorations. Our choice of t-FoR taxonomies \cite{evans2013temporal} is based on the integration of notion of transience which complements prior t-FoR taxonomies, but there is no common agreement of selecting which temporal frame of reference taxonomies to describe relations between events.

\bibliography{custom}

\appendix

% \section{Example Appendix}
% \label{sec:appendix}

\end{document}